# A Real-time Junk Food Recognition System based on Machine Learning

Sirajum Munira Shifat[1,*], Takitazwar Parthib[2], Sabikunnahar Talukder Pyaasa[3], Nila Maitra Chaity[4], Niloy Kumar[5], and Md. Kishor Morol[6]

Department of Computer Science, American International University-Bangladesh
*Corresponding author(s). E-mail(s):munirasirajum13@gmail.com [1,*], kishor@aiub.edu [6], Contributing authors: takitazwarparthib@gmail.com [2], pyaasa15@gmail.com[3],chaity13x@gmail.com [4],niloyghosh6351@gmail.com [5]

**Abstract.** As a result of bad eating habits, humanity may be destroyed. People are constantly on the lookout for tasty foods, with junk foods being the most common source. As a consequence, our eating patterns are shifting, and we're gravitating toward junk food more than ever, which is bad for our health and increases our risk of acquiring health problems. Machine learning principles are applied in every aspect of our lives, and one of them is object recognition via image processing. However, because foods vary in nature, this procedure is crucial, and traditional methods like ANN, SVM, KNN, PLS etc., will result in a low accuracy rate. All of these issues were defeated by the Deep Neural Network. In this work, we created a fresh dataset of 10,000 data points from 20 junk food classifications to try to recognize junk foods. All of the data in the data set was gathered using the Google search engine, which is thought to be one-of-a-kind in every way. The goal was achieved using Convolution Neural Network (CNN) technology, which is well-known for image processing. We achieved a 98.05% accuracy rate throughout the research, which was satisfactory. In addition, we conducted a test based on a real-life event, and the outcome was extraordinary. Our goal is to advance this research to the next level, so that it may be applied to a future study. Our ultimate goal is to create a system that would encourage people to avoid eating junk food and to be health-conscious.

## 1 Introduction

The Machine Learning approach for perceiving food is a significant recognizable factor nowadays. Numerous scientists applied this methodology in their experiments, while a portion of the investigation made noticeable progress. Recently, deep learning has been used in image recognition. The most distinguishing aspect is that better image features for recognition are extracted automatically through training. One of the approaches that meet the criteria of the deep learning approach is the convolutional neural network (CNN) [1]. Image recognition of food products is usually frequently challenging due to the wide variety of forms of food. Deep learning, on the other hand, has recently been proved to be a highly successful image recognition tool, and CNN is a state-of-the-art deep learning solution. Using computer vision and AI approaches, researchers have been focusing on automatically detecting foods and associated dietary data from images recorded over the past two decades. The main objective of this research is to recognize junk foods. Junk food is one of the popular variants in food. Michael Jacobson, who was a director of the Center for Science the expressed his feeling by saying that junk food can be considered as a dialect for foods of inadequate or has low beneficial value for the human body which is also considered as HFSS (High fat, sugar, or salt) [2].

The lack of a rapid and accurate junk food detector in the medical field has yet to be addressed, so a new deep learning-based food image recognition system is implemented at the end of this study. The suggested method uses a Convolutional Neural Network (CNN) with a few key improvements. The effectiveness of our solution has been proved by the outcomes of using the suggested approach on two real-world datasets. In our study, we have used CNN for the detection of images of junk food where we have applied YOLO technology to train our manually created datasets. A system is created to identify specific junk food using YOLO or the "You Only Look Once," framework. The YOLO model comes from the family of Fully Convolutional Neural Networks (FCN) by which the best available result can be achieved where real-time object detection is possible with every single end-to-end model.

In the current period, Food acknowledgment has stood out enough to be noticed for holding different concerns. Junk food has certainly carved the Third World for globalization [3]. It is an integral part of the lives of developed and developing countries, and the problems associated with obesity are greatly increased. Our goal is to take this research to the next level so that it can be applied to another study in the future. Our long-term objective is to develop a system that encourages individuals to eat less junk food and to be more health-conscious. This study has been conducted by following the Formal Method and Experimental Method, which will be followed throughout the whole process. Food identification and nutritional assessment studies in recent years, they have been steadily improving to establish a good dietary assessment system. Several approaches for feature extraction and categorization as well as deep learning methods have been presented.

## 2   LITERATURE REVIEW

Junk food consumption and its effects on health are becoming a major concern around the world. Everyone should receive health education so that they can recognize which foods are more helpful. Healthy eating is one of the most important tasks that will always result in a healthy lifestyle [4]. To prevent illness diet is one of the key factors [5]. Since 1966 consumption of fast food is growing at a rapid speed [6]. Any food that is simple to make and consume in a short amount of time is considered fast food. Fast food, on the other hand, lacks the necessary nourishment for human health, which is why it is less expensive. In the long run, fast food has many damaging effects on the human body due to its content of fat have excessive cholesterol levels and sugar [8]. People are now more aware of and concerned about their health. Because individuals are the main emphasis for a better future, health should be a major concern for everyone. Food identification is gaining popularity these days, like computers and mobile phones can readily detect food and its dietary fiber content using a machine learning approach [9]. In the real-time junk food image detecting process, optical sensors are used. Using optical sensors after capturing the food images every image can be labeled [10]. The work in this research is to detect junk food using the YOLOv3 algorithm detector with custom data set. The full form of YOLOv3 is 'You Only Look Once' and it is one of the fastest and commonly used deep learning-based object detection methods. It uses the k-means cluster technique to evaluate the initial width and height of the foretold bounding boxes [11] [12]. Numerous algorithms were proposed to make food recognition fast and precise. A lot of people are still working on Food Detection to make wide-ranging use of deep learning. In 2005, N.Dalal introduced Histogram of oriented gradients (HOG).Essentially, this approach is a feature descriptor that is utilized in image processing and other computer vision methods to differentiate objects. An experiment by [21] on multiple food image detection using HOG obtained 55.8% classification rate, which enhanced the standard result in case of using only DPM by 14.3 points. CNN, which means Convolutional Neural Networks is one of the most advanced methods for image recognition [13]. This technique contains a multi-layer of a neural network similar to the function of the brain [34]. The neurons take minor areas of the previous layer as input. CNN includes two layers. As a result, CNN achieved much improvement than other traditional approaches using handcrafted structures. Finished reflection of trained convolution kernels, established that color features are vital to food image recognition. Overall the output was prosperous that specifies that it would have been difiicult for humans to recognize such images [14] [15].

Mask R-CNN was used for Chinese food image detection by Y. Li, X. Xu, and C. Yuan in their paper named Enhanced Mask R-CNN for Chinese Food Image Detection. To reach their aim they first built Chinese food images. A deep separable convolution as an alternative of traditional convolution for decreasing model consumption which can significantly decrease the resource consumption and increase the detection productivity of the Chinese food images with great accuracy [18].

The Single Shot Detector (SSD) method uses a single deep neural network to recognize objects in photos. SSD eliminates the proposal generation and subsequent pixel or feature resampling stages, allowing all calculations to be consolidated into a single network. It is a direct and simple method to assimilate into systems that have need of a detection module. In both training and meddling SDD is has a great speed [30].Spatial Pyramid Pooling (SPP-net) is a network assembly that can produce a fixed-length demonstration irrespective of pictures size and scale. It is a strong method that recovers altogether CNN-based image detection approaches. It can receive both input image of different size and fixed-length piece demonstration [17].

The most efiicient and precise algorithm of all is YOLO (You Only Look Once) proposed by Joseph Redmon. It can detect multiple object from one image. YOLOv3 is very robust and has nearly on par with RetinaNet and far above the SSD alternatives. The recommended method decreases the layer number in the current YOLOv3 network model to shorten the failures and feature mining time [31].A deep convolutional neural network is combined with YOLO, a contemporary object identification approach, to identify two objects at the same time. The accuracy rate was around 80% [16].

Centered on Faster R-CNN with Zeiler and Fergus model (ZF-net) and Caffe framework. p-Faster R-CNN algorithm for healthy diet detection was projected. By comparing it with Fast R-CNN and Faster R-CNN the performance of p-Faster R-CNN was weighed. Faster R-CNN rises the AP value of each kind of food by more than 2% paralleled with Faster R-CNN, and p-Faster R-CNN, Faster R-CNN is superior to Fast R-CNN correctness and speed [17] [18].

To test the proposed algorithm's accuracy and speed against current YOLOv3 and YOLOv3-tiny network models using WIDER FACE and datasets, it was compared to existing YOLOv3 and YOLOv3-tiny network models. As a consequence, the accuracy performance comparison results for YOLOv3 were 88.99%, 67.93% for YOLOv3-tiny, and 87.48% for YOLOv3-tiny. When compared to the present YOLOv3-tiny, this result reveals a 19.55% improvement accuracy. In comparison to YOLOv3, the outcome was 1.51% lower. In addition, the speed was 54.7 FPS lower than YOLOv3-tiny. On the other hand, 70.2 frames per second are quicker than YOLOv3 [19] [20].

When a person looks at an image they can immediately determine the object, notice their position and even know how to react to that. A human can detect an object precisely even if the image is blurred or in complex structure [21]. However, the difiiculty is that the globe is vast, and not everyone is aware of the variety of junk food available. Junk food is different in different countries. With the use of a prediction model, the system will alert the user to the food's effect. The majority of people have food allergies. One of the most prevalent scenarios is an adverse reaction. People can be more conscious of their meal choices by detecting these types of food [22]. Contemporary object detection approaches, such as the CNN family and YOLO, are briefly discussed in this work. In actuality, YOLO has a wider use than CNNs. A unified object detection model is the YOLO object detection model. It's simple to put together and can be quickly trained on whole photographs. Unlike classifier-based approaches, YOLO is trained on a loss function that is directly related to detection. Furthermore, because YOLO provides a more generalizing representation of objects than other models, it is well suited to applications requiring speedy and accurate object detection [23].

## 3  METHOD

The research method is selected based on a machine learning approach with 20 different classes of various junk food.The process is followed in machine learning problem solving technique. YOLO (You Only Look Once) object detector is chosen since the detector is rapid and reliable at the same time. The detector configuration was tuned to get the most efiicient outcome of 98.05% accuracy.The following workflow are followed for the experiment:

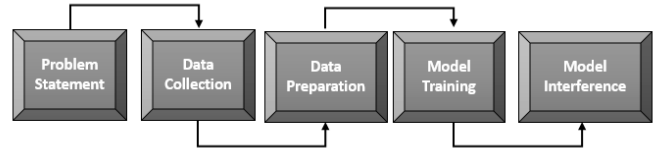

Fig. 1. Research Workflow

### 3.1  About Model (YOLOV3)

The YOLO model comes from the family of Fully Convolutional Neural Networks (FCN) by which best available result can be achieved where real time object detection is possible with every single end-to-end model. In YOLO object detection model, the train image is divided into several grid (13x13). On these grids 5 bounding boxes are created [32]. One bounding box may intersect several grids as well. Confidence level is assigned for each bounding box. With a certain threshold the bounding boxes are selected for classification. The model can already recognize 15 objects. 20 more items were manually added with the given model. 100 Pictures were selected for each class. Then with the help of data augmentation method the dataset was increased to around 500 Pictures for individual class. Later, the images were labeled with YOLO format using "labelImg".

It has 75 convolutional layers. Due to the Residual layer, it can contain skip connections. It has upsampling layers as well. There is no use of pooling in YOLO v3. To avoid loss of low-level features stride 2 is being used in each convolutional layer. YOLO can handle different shape of input images but if the shapes are similar it makes it less complicated to debug. After the convolution a feature map is generated which contains (B x (5+C)) entries (where B is number of bounding boxes and C is the class confidence for specific bounding box) [32]. Each bounding box has 5+C attributes- (x, y) co-ordinates, w (width), h (height), objectiveness score and C class confidences. The specialty of YOLO v3 is, it can predict 3 bounding boxes (Anchors) for individual cell. Anchors are preset bounding boxes that have been pre-defined. The backbone of YOLOv3 is Darknet-53. Other models like SSD and RetinaNet, on the other hand, use the common ResNet family of backbones [33]. Also, YOLOv3 is an incremental upgrade from previous models. It preserves fine-grained features by sampling and concatenating feature layers with earlier feature layers. Another enhancement is the use of three detection scales. As a consequence, the model can recognize objects of varying sizes in images. The link of the process of the whole algorithm is given:
https://github.com/MasterS007/YOLOV3-Process-for-detecting-Junk-Food.git

### 3.2 Dataset Collection

Dataset was collected manually for 20 different classes of junk food such as biryani, burger, chicken nuggets, hotdog, chips, Kabab, Mac and Cheese, Meatloaf, Mufiin, Nachos, Cookies, French Fry, Ice-cream, Pizza, Processed Cheese, cheesecake, crispy chicken, sandwich, noodles and waffles (Figure 2). The pictures were captured by smartphones and Digital Cameras. Foods from restaurants, food courts, and Cart parked on street were taken into consideration to add variety to the dataset. The pictures were searched from internet. The most popular pictures were taken from Google Images. Along with that some social media (Facebook) was used as well to get latest pictures of the classified foods.100 images were selected per class resulting in 100x20=2000 images in total. Link of the dataset is:
https://www.kaggle.com/sirajummunirashifat/junk-food-recognition-system-based-on-ml

### 3.3 Data Augmentation

Since the number of classes was comparatively high, a simple test was conducted by training with initial 2000 images. The training results with such a low amount of dataset only provided 61.6% accuracy. Data augmentation was used to further increase the dataset since a low dataset might cause low accuracy. Rotation of random degrees, horizontal and vertical flip, zoom in rescaling was taken into consideration for augmentation. This was used so that the variety of the data can be trained with the same image. After the augmentation, there were 500 images for each class. Which resulted in 500x20=10000 images.

### 3.4 Data Preparation

The pictures were all in various formats so they had to be resized to a 416X416 and renamed in a certain order to reduce complexity. All the pictures were converted to jpg for YOLO bounding box creation. All the 10000 images were labeled manually. The open-source software labelImg [24] also supported multiple bounding boxes in a single picture which increased the number of total bounding box helping the accuracy of the training. Coordinates(Tx, Ty, Th, Tw) and the class number of the object were extracted from each image by labeling to accommodate it for training.

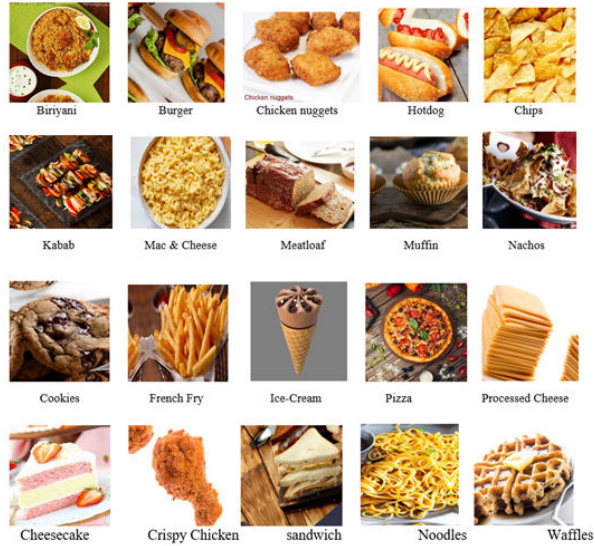

Fig. 2. Sample Images of Classes

### 3.5  Model Tuning

The working mechanism of the object detection model is based on custom configuration YOLO v3. Darknet-53 was used as a base to start the tuning. The detection is completed with 3 phases of detection in the RGB channel. Inputs were taken as 416x416 to save training time. The model uses 9 anchor boxes generated with K-Means clustering. 75 Convolution Layers are executed with filters size varied from 32, 64, 128, 256 and max up to 512. The filter size was increased to get more accurate results while sacrificing very little detection time. Stride and Padding 1, 2 is used for all convolution layers with Leaky Relu (equation 2) as an activation function. Binary cross-entropy loss is used as an error function to avoid overlapping classes. Simple Logistic regression is applied to measure objectness score. Tensor number 75 x 75 x [3*(4+1+80)] were calculated from (equation 1)  for 80 classes.

$$T = S.S.B(4 + 1 + C) \tag{1}$$

Here 75 x 75 kernel size was inserted in the 3 detection layers. The higher layers are responsible for the smaller case of the object whereas the lower layer (82nd) is accurate with larger object enabling detection across all scaling. SGD optimizer is applied with a learning rate of 0.001 for adaptive learning. Many models were trained to tune in for the best outcome. Models kernel size for each convolution was increased to notice the performance increase. Batch normalization was applied with testing on subdivisions 16 and 32.

$$f(x) = 1(x < 0)(0.1x) + 1(x >= 0)(x) \tag{2}$$

### 3.6   Internal Architecture

Approaches for each epoch that have been applied to train the model using Darknet-53 are shown in Figure 3. The batch size was taken 64 and subdivision was taken 16. Max batch was 10000, used SGD optimizer.

```
For each epoch:
Batch:64, Subdivision:16, Max Batch: 10000
Optimizer: SGD (Learning Rate:0.001)
```
| |
|---|
| 4x Convolution (2xFilter 64x64, 2x filter 64x64) |
| Shortcut Connection |
| 3x Convolution (1xFilter 64x64, 2x filter 128x128) |
| Shortcut Connection |
| 2 x Convolution (1xFilter 64x64, 2x filter 128x128) |
| Shortcut Connection |
| 3 x Convolution (1xFilter 128x128, 2x filter 256x256) |
| Shortcut Connection |
| 7 x [2x Convolution (1xFilter 128x128, 1x filter 256x256) Shortcut Connection] |
| 3 x Convolution (2x filter 256x256, 1x filter 512x512) |
| Shortcut Connection |
| 7 x [2x Convolution (1x filter 256x256, 1x filter 512x512) Shortcut Connection] |
| 3 x Convolution (1x filter 512x512, 2x filter 1024x1024) |
| Shortcut Connection |
| 6 x [ 2x Convolution (1x filter 512x512, 1x filter 1024x1024) Shortcut Connection] |
| Convolution (filter 75x75) |
| 1$^{st}$ Prediction Layer |
| Convolution (filter 256x 256) |
| 3 x [ 2x Convolution (1x filter 256x256, 1x filter 512x512) Shortcut Connection] |
| Convolution (filter 75x75) |
| 2$^{nd}$ Prediction Layer |
| 4 x [ 2x Convolution (1x filter 128x128, 1x filter 256x256) Shortcut Connection] |
| Convolution (filter 75x75) |
| 3$^{rd}$ Prediction Layer |

Fig. 3. Internal Architecture of the model

### 3.7   Model Training

The model was conducted with an Nvidia RTX 2060 GPU and Google Colab using the python Tensorflow library. The training data was split into training, validating, and testing sections. 100 Epochs were selected for each iteration. A total of 80% of the data was chosen for training, while the remaining 20% was retained for testing and validation. A total of 40000 epochs were iterated with the required time of 3 days. Each iteration kept decreasing the neat loss. The weights were generated after 100 epochs which could be used on any platform for testing.

## 4 MODEL EVALUATION

### 4.1 Training and Testing

The experimental findings of our system employing the dataset of junk food of 20 classes are shown in this chapter where each item consists of 500 images, which we created ourselves. To recognize junk food, the Yolov3 algorithm was proposed and the results were more accurate. We needed to move this properly structured dataset into VM (Virtual Machine) cloud because Google Colab was used to train this model. After moving the dataset into the VM cloud the yolov3-custom.cfg file was configured. To configure the file, the batch size was changed to 64, and the subdivision was changed to 16. We set our max-batches to 40000, steps to 32000, and 36000, the class value was altered to 20 in all three YOLO layers, and filters were set to 75 in all three convolutional layers before the YOLO layers.

### 4.2 Model Performance

After successfully training the model, we made a quick test with some junk food images captured by our cellphone. The model was testing with unseen image dataset. (Figure 4) shows some testing results of images.

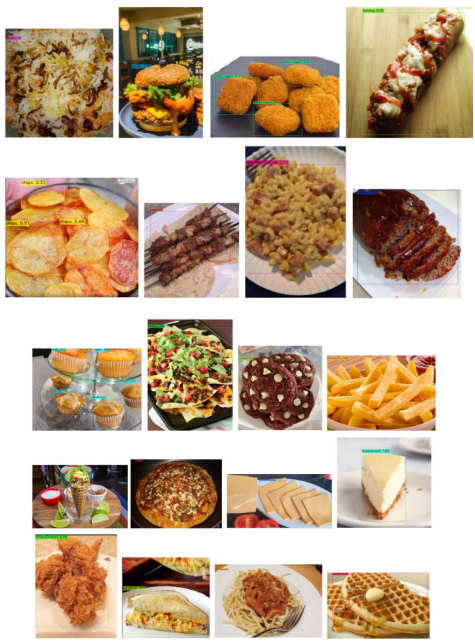

Fig. 4. Result of Food Recognition

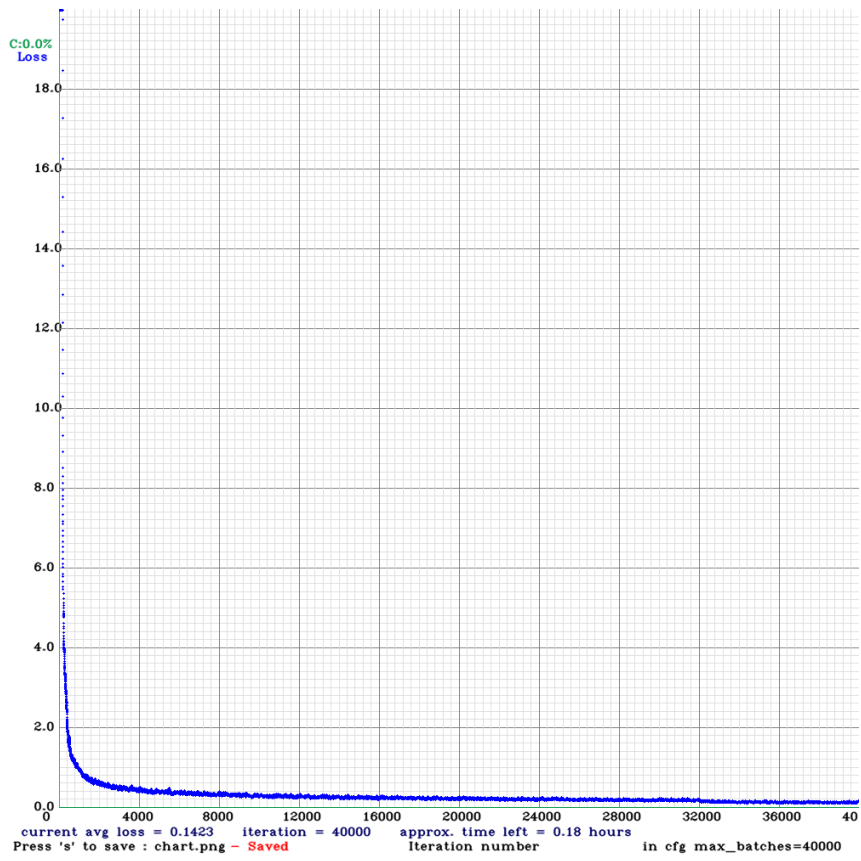

Fig. 5. Graphical view of loss of training

Our trained model has a very low loss, as shown in the graph (Figure 5). For our model to be 'accurate' we would aim for a loss below 2. In this graph, we can see that the model's loss is less than 2, indicating that we achieved our projected aim. For the training dataset, our suggested model has an accuracy of 98.05%. The exam on random photos went reasonably well when the training was completed. We succeeded to achieve a pretty consistent pace.

|  | Predicted duration | | Percentage of Error | | Percentage of Accuracy | |
|---|---|---|---|---|---|---|
|  | Colab GPU with subdivision 16 | Local GPU with subdivision 32 | Colab GPU with subdivision 16 | Local GPU with subdivision 32 | Colab GPU with subdivision 16 | Local GPU with subdivision 32 |
| Total Average | 40.34ms | 40.33ms | 1.95% | 12.8% | 98.05% | 87% |

Fig. 6. Accuracy of Food Recognition from testing images

In Figure 6 it is shown that the test image dataset was used in the model which trained to test the accuracy in Google Colab with 16 number of subdivision and also used in the model which trained in local GPU machine with 32 number of subdivision.

With the help of image augmentation, a dataset of 10000 images with 3 days of training was able to produce results with upwards of 98.05% confidence. Given the complexity of the food images, the model did exceptionally well. We also trained the model on a local GPU system, changing the batch size and subdivisions to 64 and 32, respectively. The major motivation for this was to make subdivision 32 so that training the model would take less time. If the subdivision number is higher, the amount of time spent training will be reduced. As a result, the model's subdivision number was increased. Even though training time was reduced and training accuracy was excellent, test accuracy was dismal. We can see from the above comparison that when the subdivision was 16, the model was more accurate 98.05% confidence, and when the subdivision was 32, the model was less accurate 87% confidence.

### 4.3  Result Comparison with Previous Research

A sliding window was used in conventional computer vision systems to scan for things at different locations and sizes. The aspect ratio of the item was typically thought to be fixed because this was such a costly task. Selective Search was employed by early Deep Learning-based object detection algorithms like the R-CNN and Fast R-CNN to reduce the number of bounding boxes that the algorithm had to examine. Another method, termed Overfeat, included scanning the picture at various scales using convolutional sliding windows-like techniques. Faster R-CNN was then used to identify bounding boxes that were to be tested using a Region Proposal Network (RPN). The traits retrieved for object recog-nition were also employed by the RPN for proposing probable bounding boxes, which saved a lot of work.

YOLO, on the other hand, takes an entirely new approach to the object detection problem. It merely sends the entire image via the network once. SSD is another object detection approach that passes the picture through a deep learning network once, although YOLOv3 is significantly quicker and achieves similar accuracy. On an M40, TitanX, or 1080 Ti GPU, YOLOv3 produces results that are quicker than real-time with more accuracy.

Yolo Implementation model is like a normal CNN model. There are many CNN based algorithm like RCNN, Yolo model and many more. But R-CNN algorithm requires thousands of network evolution to predict one image which are very time consuming. R-CNN algorithm focuses on specific area of the image and trains separately each component. To solve this problem YOLO model has been proposed. The main purpose of use this algorithm is, it is computationally very fast and also used on real time environment.

It is difiicult to find research that identifies this type of junk food. However, there are a few papers that are comparable to this one. The table below (Figure

| Work | Technology Used | Accuracy |
| --- | --- | --- |
| Food image recognition using deep convolutional network with pre-training and fine-tuning [25]. | DCNN | 70.4% |
| Im2calories: towards an automated mobile vision food diary [26]. | Google net/food101 | 79.0% |
| A food recognition system for diabetic patients based on an optimized bag-of-features model [27]. | ANNnh | 75.0% |
| Yelp Food Identification via Image Feature Extraction and Classification [28]. | CNN & SVM | 68.49% |
| Food recognition: a new dataset, experiments and results [29]. | CNN | 79.0% |
| **A Real-time Junk Food Recognition System based on Machine Learning.** | **CNN with YOLOV3** | **98.05%** |

Fig. 7. Result comparison with previous research

7) shows some of the connections between our research and previous studies on food classification.

## 5 Conclusion

In this study, A CNN-based algorithm was implemented by modifying Darknet-53 architecture. The model was heavily tuned to achieve an outcome of 98.05% accuracy. Our model was trained in The dataset contains 20 classes and 500 images were taken for each class. It's evident that our object detection model outperforms other categorization methods. The model provided almost 30 FPS on a real-time windows desktop app on an Intel Core I5-7400 desktop. The generated weights can be used on Android, IOS devices to create a portable application. With a single snap, people will be able to capture and detect the desired food. In the future, our team intends to develop such an app. Though the testing was conducted on a small test set of 1000 images, further testing is expected to keep an average accuracy of more than 90%. In this research, we also introduced a dataset of 10000 images of junk food which was captured and gathered by us. The limitation we have faced while training our model is a lack of resources such as a strong enough GPU to reduce the training time.

Because junk food is linked to a variety of severe diseases, we will be able to utilize our model to determine the link between those diseases and junk food in the future. With the use of this technology, new research in the medical and technological fields can be undertaken so that people can avoid eating foods that are detrimental to them.